\documentclass{article}

\usepackage[preprint, nonatbib]{neurips_2022}

\usepackage[utf8]{inputenc} 
\usepackage[T1]{fontenc}    
\usepackage{hyperref}       
\usepackage{url}            
\usepackage{booktabs}       
\usepackage{amsfonts}       
\usepackage{nicefrac}       
\usepackage{microtype}      
\usepackage{xcolor}         
\usepackage{graphicx}
\usepackage{authblk}
\usepackage{amsmath}

\title{Proactive Detractor Detection Framework Based on Message-Wise Sentiment Analysis Over Customer Support Interactions}

\author[]{J. S. Salcedo-Gallo}
\author[]{J. Solano}
\author[]{H. García}
\author[1, 2]{D. Zarruk-Valencia}
\author[]{A. Correa-Bahnsen}
\affil{Rappi AI Research}
{
    \affil[1]{\{sebastian.salcedo, jesus.solano, javier.garcia, alejandro.correa\}@rappi.com}
    \affil[2]{davidzarruk@gmail.com}
}

\begin{document}

\maketitle

\begin{abstract}
In this work, we propose a framework relying solely on chat-based customer support (CS) interactions for predicting the recommendation decision of individual users. For our case study, we analyzed a total number of 16.4k users and 48.7k customer support conversations within the financial vertical of a large e-commerce company in Latin America. Consequently, our main contributions and objectives are to use Natural Language Processing (NLP) to assess and predict the recommendation behavior where, in addition to using static sentiment analysis, we exploit the predictive power of each user's sentiment dynamics. Our results show that, with respective feature interpretability, it is possible to predict the likelihood of a user to recommend a product or service, based solely on the message-wise sentiment evolution of their CS conversations in a fully automated way.

\end{abstract}


\section{Introduction}
\label{section:introduction}

Since the introduction of the Net Promoter Score (NPS) \cite{Reichheld2003} in the early 2000s, it has become an important metric for measuring customer recommendation behavior across multiple industries \cite{Raassens2017, Ando2017, Bockhorst2017, Siering2018, Auguste2018, Chatterjee2019, Rose2019, Vanderheyden2019, Lewis2020, Markoulidakis2020, Baehre2021, Jain2021, Markoulidakis2021, Zaki2016}.
This score is based on a rather straightforward question, namely \textit{'How likely would you be to recommend us to your friends or family?'} Moreover, these surveys are massively conducted at a frequency ranging from monthly to an annual basis \cite{Markoulidakis2020}, or even immediately after individual interaction with Customer Support (CS) \cite{Auguste2018}. In practice, the NPS score can be segmented into three main groups, namely \textit{promoters}, \textit{passives}, and \textit{detractors} \cite{Reichheld2003}. 


\begin{figure}
  \centering
  \includegraphics[width=0.8\textwidth]{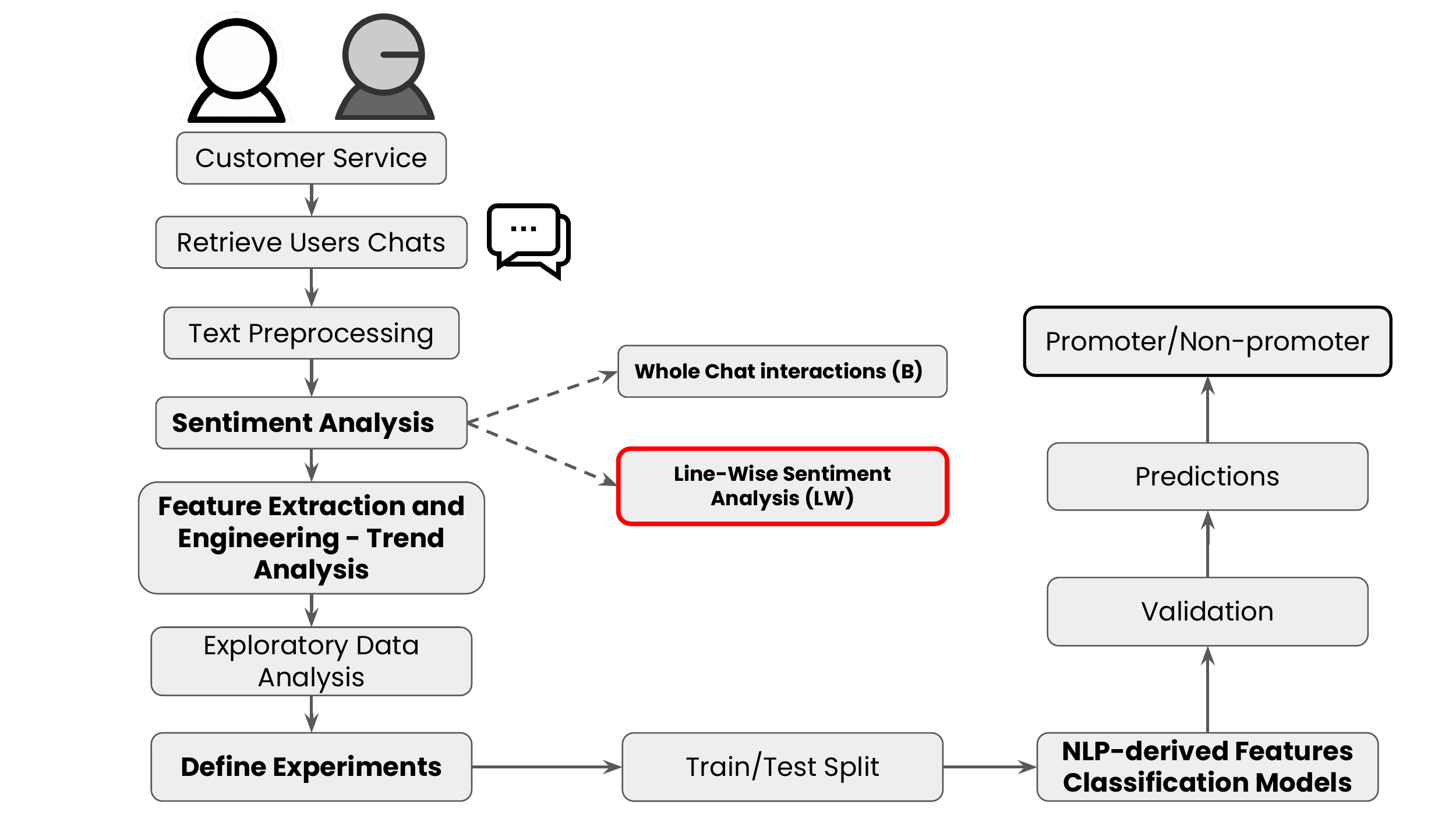}
  \caption{Our proposed framework for proactive \textit{promoter}/\textit{non-promoter} user detection for any chat-based user interaction.}
  \label{fig:pipeline}
\end{figure}

Previous studies on sentiment analysis related to NPS recommendation behavior have not been studied in chat-based customer relationship interactions, which have gained popularity given the massive outreach of widely adopted messaging applications and Client Resource Management (CRM) platforms. Furthermore, it is of compelling interest for companies to predict recommendation decisions to proactively and effectively correct service failures, and retribute customers with incentives, which ultimately could improve the company's ratings. Therefore, our main contribution and objective are to use NLP to assess and predict the recommendation behavior of the users where, in addition to using static sentiment analysis, we exploit each user's sentiment evolution throughout a particular conversation, inspired by the inherently dynamic nature of human conversations \cite{Garas2012}, while employing widely-known and robust ML transformed-based architectures \cite{BERT2018}.

In this work, we address the NPS binary classification problem, namely predicting whether a given user would be a \textit{promoter} or not. We have a total number of 16.4k users and 48.7k CS conversations with a large e-commerce company in Latin-America for one specific city, where each particular recommendation rating is obtained directly from the users themselves in corresponding survey responses. With this, we have performed sentiment classification based on transformer-based architecture over each user's CS chat-based interactions to predict well in advance the recommendation behavior of particular users. Each turn-level inherent sentiment evolution feature is used to perform a classification task at hand. We draw inspiration from existing approaches for conversational discourse-aware response generation such as DialogBERT \cite{Gu2021}, DialoGPT \cite{Zhang2019}, or Blender \cite{Roller2020} for our analysis.


Furthermore, we propose a message-wise sentiment evolution analysis of the customer-sent messages throughout the conversation. This message-wise approach consists of obtaining a sentiment score for each message the user sends in a particular chat-based CS conversation. Thus, allowing us to analyze the trend and overall sentiment evolution and their relation with the recommendation behavior of any particular user. Our results show that overall, message-wise evolution analysis is substantially superior in predicting the recommendation behavior of individual users compared to a traditional review-based approach of computing the overall sentiment of the complete interactions. To the best of our knowledge, this is the first message-wise sentiment evolution analysis over chat-based CS interactions to predict the recommendation behavior of users, which is highly extensible for any particular domain and any specific business-driven metric. For instance, N-class or binary problems, such as the Satisfaction Score (CSAT), Churn score, Recency-Frequency-Monetary (RFM) score, or even fraud detection, to name a few. 


\section{Background}
\label{section:background}

The most relevant advances in algorithms to predict the NPS score are based on online reviews \cite{Siering2018}, word of mouth (WOM) \cite{Raassens2017}, social media data analytics \cite{Vidya2015}, explicit quantitative experience attributes and service ratings \cite{Markoulidakis2020, Markoulidakis2021}, among others\cite{Jain2021}. These previous approaches used simple regression models \cite{velez2020}, tree-based classification approaches such as simple Decision Trees \cite{Markoulidakis2020}, as well as Support Vector Machines \cite{Markoulidakis2020}, Deep Neural Networks \cite{Auguste2018}, and probabilistic approaches \cite{Markoulidakis2021}. Moreover, for obtaining quantitative insights from text, there is evidence of manually and empirically performed analysis based on explicit human annotations to assess the negative, neutral, and positive polarity of eWOM messages, and their correlation with detractors, passives, and promoters recommendation decisions \cite{Raassens2017}. Also, other classification approaches for `\textit{Recommend}' or `\textit{No-recommend}' behavior have considered more robust modeling such as bag-of-words and aspect-based sentiment analysis over online airline reviews \cite{Chatterjee2019, Siering2018}. However, on one hand, the actual predictive powers of combining qualitative and quantitative user-generated content on the recommendation behavior were not explored and are limited to correlation and statistical analysis so far. On the other hand, in the case of online reviews, where authors \cite{Siering2018} used overall-sentiment, aspect-specific-sentiment, and bag-of-words, they indeed address the performance on the prediction task of recommending/non-recommending. However, their results and approach are not generally comparable with ours, given the fundamental difference between chat-based interactions and online reviews \cite{Siering2018, Markoulidakis2020, Auguste2018, Bockhorst2017, Baehre2021}. Consequently, the actual predictive powers of inherent characteristics obtained from the sentiment evolution in chat-based interactions were yet to be explored and assessed in general, and in a real-world environment.

An earlier study \cite{Chatterjee2019} showed that positive emotions have a positive relationship with customer outcomes, whereas negative emotions had a negative relation with recommendation behavior regardless of the nature of the service. As such, the author made relevant contributions from a theoretical point of view. However, the actual predictive powers on a classification task were not addressed nor explored. 


Finally, authors in \cite{Jain2021} stated that, for future work, ensembles and optimization procedures should be applied for getting predictive recommendation decisions in the scope of the NPS prediction problem. As such, we here explore the use of an ML Ensemble model, more specifically Gradient Boosting Trees. All of this, in the light of a real-life dataset, while studying the impact of using a random under-sampling technique for training in order to improve generalization.

\section{Detractor Detection Framework Based on Sentiment Analysis}
\label{section:methodology}

In this work, we propose a framework for proactive detection of the recommendation behavior in the scope of the NPS classification problem considering the user's chat-based CS interactions, as shown in Figure \ref{fig:pipeline}. The main idea lies in the hypothesis that the trend and sentiment evolution of chat-based conversations can better capture the overall behavior of the customers toward their recommendation decisions. Our method consists on classifying the sentiment of each message with language modeling based on ML transformers architecture. For instance, let us consider a user that started a conversation rather neutral, then swung to rather negative consecutive messages while the issue is being exposed, to then end the conversation rather positively as the issue might have been effectively solved. In this case, the user is presumably more likely to recommend the service as a result of good customer experience or low frustration. 


\subsection{Sentiment analysis of CS conversations}
\label{subsection:sentiment-analysis-cs-conv}

Specifically, for the sentiment measure $SS(\cdot)$ we use a sentiment classifier based on the Bidirectional Encoder Representations from Transformers (BERT) architecture \cite{BERT2018}, which is a transformers-based ML technique for NLP applications such as sentiment classification \cite{Munikar2019}, question answering \cite{Zaib2020}, masked language models\cite{Bao2020}, Next Sentence prediction tasks \cite{shi2019}, among others \cite{Liu2019RoBERTaAR}. In this case, we used a specific pre-trained model that is fine-tuned and intended for direct use as a sentiment classification model for product reviews in English, Dutch, German, French, Spanish, or Italian. Regardless of the language, this model has a sufficiently robust off-by-one accuracy ranging from 93\% to 95\%, corresponding to the percentage of reviews where the number of stars the model predicts differs by a maximum of 1 from the number given by the human reviewer in a 0-4 scale.


Once we obtain the message-wise sentiment per user's responses in each specific longest conversation, we perform manual feature extraction and feature engineering based on the retrieved sentiment classification. For the conversation $\hat{c}\in C_i$ we construct the {\bf discrete sentiment line} as the curve associated to the vector $\text{disc} = MWS(\hat{c})$. More precisely, for a given message $m$ the model return the number of stars associated to $m$ as the value $SS(m)$. This model also returns the probability $\mathbb{P}(SS(m))$. Given that this `discrete' approach ($MWS(\hat{c})$) might result in a very step-like behavior in general, we therefore opted for having a `continuous' sentiment, namely the actual star having the largest probability score plus the actual score associated with that particular star. Formally, we define the {\bf continuous sentiment curve } as the curve associated to the vector:  
\begin{equation}
\text{cont}(\hat{c}) := (SS(\hat{m}_1)+\mathbb{P}(SS(\hat{m}_1)),..,SS(\hat{m}_N)+\mathbb{P}(SS(\hat{m}_N))) 
\label{eq:continous}
\end{equation}
where $\hat{c}:=(\hat{m}_1,...,\hat{m}_N)$.


To further smooth the sentiment message-wise series, we apply an exponential weighted mean function to Eq. \ref{eq:continous}, so that we obtain a more flexible evolution onset, more suitable for trend and concavity analysis:

\begin{equation}\label{expo}
MA_j := \alpha \cdot \text{cont}(\hat{c})_j + (1-\alpha)\cdot MA_{j-1}
\end{equation} which we depict as a red dashed line in Fig. \ref{fig:line-wise messages}, where we consider an static decay parameter of $\alpha = 2/3$. We apply a simple linear regression to capture the trend, which in general can give an approximation for overall evolution of the conversation. Then this value slope is used as a feature for our classification model.

Finally, we define the experiments, namely the sampling technique for unbalanced treatment, target labels, and the classification model to train. For this, we use a widely-known ML classification algorithm, namely Gradient Boosting Trees (XGBoost) \cite{XGBoost2016} and Random Search hyper-parameter tuning for the classifier \cite{Montovani2015}. The proposed framework is very flexible as it could be easily extended to any chat-based interactions to predict any business-driven metric or N-class customer rating classification task, such as Satisfaction Score (CSAT), Churn score, Recency-Frequency-Monetary (RFM) score, or even fraud detection.


\subsection{Baseline Model}
\label{subsection:baseline-approach}

In order to assess the relevance of the here proposed framework for chat-based CS interactions, we obtain a baseline model that mimics an empirical sentiment estimation for batch text of the complete interaction. First, for our baseline model, we consider the static sentiment of each complete interaction. Let $C_i$ be the set of complete conversations of the user $u_i$. For each $c\in C_i$ we compute the static sentiment $SS(c)$ using the architecture described in section \ref{subsection:sentiment-analysis-cs-conv}. Therefore, we compute the descriptive statistics variables such as the mean, minimum, maximum, and median sentiment of the vector $(SS(c))_{c\in C_i}$, as well as the number of CS interactions,  $|C_i|$. Accordingly, with this baseline set-up features we can train a model, following procedure described in Section \ref{subsection:experimental-setting}, that gives an estimation of the probability of a given user to be a \textit{promoter} based on overall static sentiment evaluation. However, it is very relevant to make clear that our main contribution is not just to use sentiment analysis and Natural Language Processing (NLP) to predict recommendation behavior, but rather to use highly mature components \cite{BERT2018} in a novel way for capturing the actual sentiment evolution throughout the CS conversations, as presented in Sec. \ref{subsection:sentiment-analysis-cs-conv}.

\subsection{Line-wise Sentiment Analysis Model}
\label{subsection:line-wise-sentiment}
\begin{figure*}
    \centering
    \includegraphics[width=0.8\textwidth]{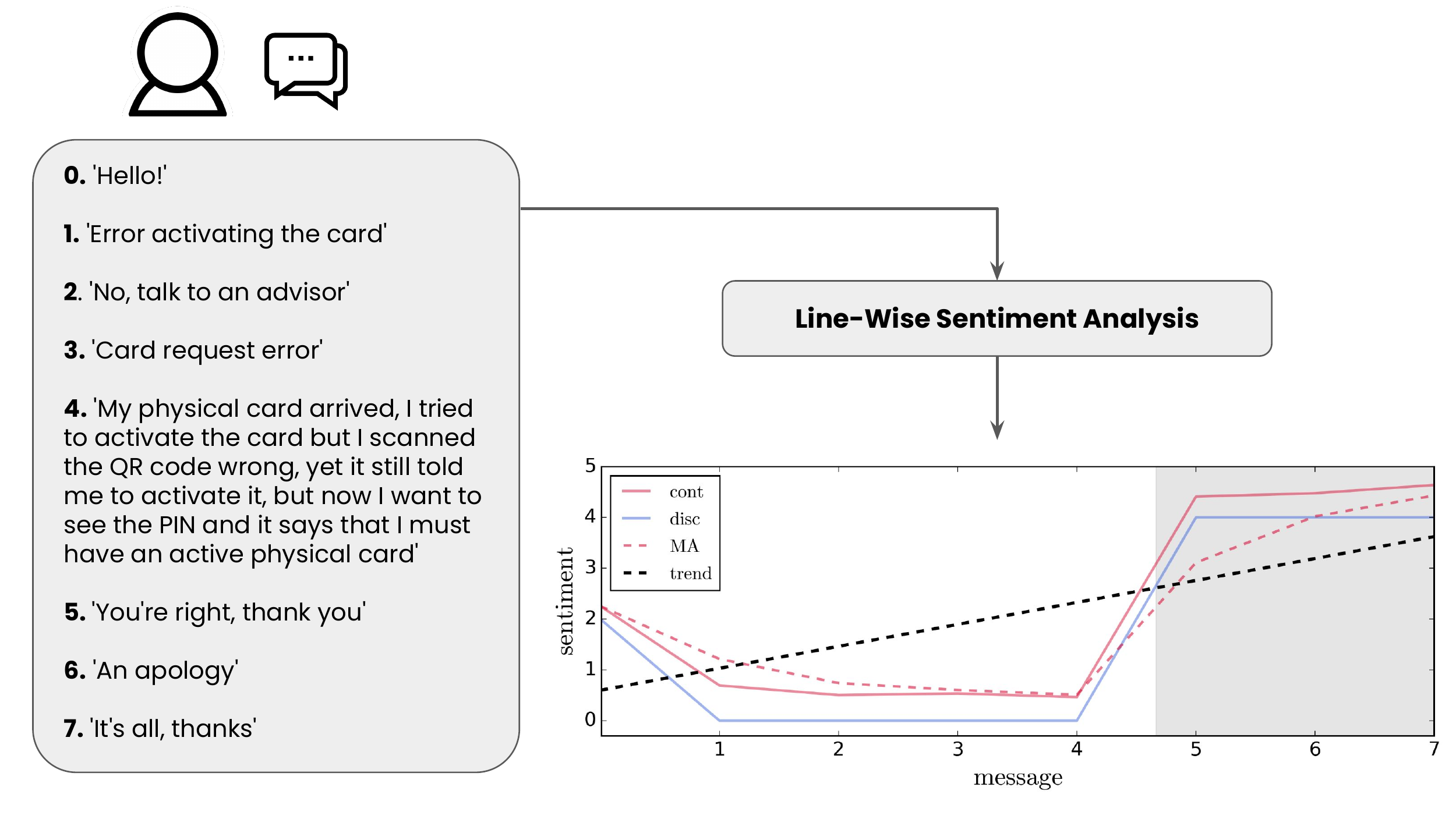}
    \caption{We depict a typical example of user-sent messages to CS in a particular interaction. The blue solid line represent the `discrete' sentiment, which corresponds to the star rating computed with the sentiment classifier per each user's message. The red solid line represent the `continuous' sentiment, which corresponds to the discrete star rating plus the actual score associated with that particular star. The red dashed line represents an exponential weighted mean function (MA), and the black dashed line represent a linear fit over the smoothed MA curve. The shaded region corresponds to the last third of the conversation, which we have used to extract some features for our analysis. We only plot the linear fit to the conversation, even though concavity and descriptive statistics features are also computed from this message-wise analysis.}
    \label{fig:line-wise messages}
\end{figure*}

We perform granular, flexible and general message-wise sentiment analysis to assess the inherent sentiment dynamic nature of conversations and how it relates to the recommendation behavior of a given user. To the best of our knowledge, there is no scientific evidence showing that the actual predictive powers of sentiment evolution in chat-based interactions have been explored and assessed regarding general binary classification tasks.


As such, we assess the predictive power of quantitative message-wise sentiment analysis features extracted from individual CS conversations. Consider a conversation $c\in C_i$ and its vector representation $c:=(m_{1},...,m_{N})$, where $m_{j}$ is the $j$-th user message or turn appearing in the conversation, thus neglecting agent responses. We begin considering the vector of message-wise sentiment as $MWS(c):=(SS(m_1),...,SS(m_N))$. Figure \ref{fig:line-wise messages} shows an example of a typical CS conversation, where each message has an associated sentiment on the message-wise panel on the right. It can be seen that the conversation started rather neutral, then there were some issues exposed in the middle part of the conversation, which were effectively solved by the agent.


Let $\hat{c}$ be the longest conversation in $C_i$. Once the message-wise sentiment vector $MWS(\hat{c})$ is computed, we obtain descriptive statistics features, as well as concavity analysis related to the mean of the numerical second derivative and the overall trend fitted as a linear slope. With this, we are able to capture not only the overall trend, but also the evolution of the conversation in terms of concavity analysis. A more detailed description of this process appears in Section \ref{subsection:sentiment-analysis-cs-conv}. We look at the evolution of the message-wise sentiment vector $MWS(C_i)$ over time. This includes both the number of messages belonging to each specific discrete sentiment class, as well as the descriptive statistics features extracted from the last third of the conversation. For the here presented framework, the overall sentiment of the complete interaction is compared against the granular sentiment evolution of the chat-based messages.


\section{Evaluation}
\label{section:evaluation}
\subsection{Dataset}



Our dataset consists of the CS conversations and NPS surveys conducted in the second semester of 2021, of users within the financial vertical of a large e-commerce company for one specific city. In our case study we consider around 16.4k users having CS interactions whom also have filled the NPS survey, for a total number of interactions to the CS center $\bigcup C_i \approx$ 48.7k+. Also, the mean number of interactions correspond to 2.39. Therefore, we predominantly base our analysis on the largest conversation of a given user, while still considering general features for the overall CS interactions. Moreover, the mean number of messages for the largest conversations per user corresponds to 13.85 messages. We use only the customer responses, thus ignoring the agent's prompts. After querying the conversations, we then apply a simple preprocessing step, namely removing special characters and blank messages. As for the count of each recommendation behavior annotated by the users themselves in our dataset, we have a total of 10701 \textit{promoter} users, whereas 5700 \textit{non-promoter} ones, making it a slightly imbalanced classification problem. It is of compelling relevance to point out that we use only features extracted from text, thus preventing ourselves from using behavioral data or other specific service ratings.

\subsection{Experimental Setup}
\label{subsection:experimental-setting}

We use Random Search for hyper-parameter tuning of the XGBoost classifier used in this work, where both the number of folds in a (Stratified)-KFold and the number of parameter settings that are sampled for each classifier were set to 10. For our experimental setup, we have a total of 16401 users, for which we use a user-wise train/test split ratio of 80\% / 20\% for the train and test set, respectively. We also explore the impact of sampling the train set over the predictive power in terms of validation metrics such as the F1-score, and Specificity \cite{Goutte2005}. Moreover, we also assess the influence of \textit{passive} (2470) users on the predictive power of the algorithm, so that we additionally obtain classification results considering only \textit{promoters} (10701) and \textit{detractors} (3230) users. In that regard, we consider three experiments in our setup, namely our (1) static-like sentiment baseline \textbf{(B)}, (2) message-wise sentiment evolution analysis considering \textit{passive} users \textbf{(B + LW)} , and (3) message-wise sentiment evolution ignoring \textit{passive} users \textbf{(B + LW\{NP\})}.


We then validate the performance of the classification models for the different experiments using the Area Under the Curve (AUC) \cite{Hanley1982}, Kolmogorov-Smirnov (KS) metric \cite{KS2008}, and Macro F1 score \cite{Goutte2005} as our validation metrics for one-to-one comparison. 

\section{Results and Discussion}
\label{section:results}


Table \ref{tab:table1} shows the results of our classification task, namely predicting whether the given customer would be a \textit{promoter} (1) or not (0). Results are presented by considering Random Undersampling only over our train set for dealing with imbalanced nature of our dataset. We show results for our baseline \textbf{(B)} approach, as well as for our dynamic line-wise sentiment analysis considering \textbf{(B + LW)} and ignoring \textbf{(B + LW\{NP\})} \textit{passives} users. 


First, the results presented in Table \ref{tab:table1} for our baseline \textbf{(B)} show that this model is only marginally doing better than a random guesser. Therefore, the overall and static sentiment of complete CS interactions does not present strong predictive power considering only CS chat-based interactions. Nonetheless, they do have comparable performance if compared against to previous works where explicit service ratings were used as features, exhibiting a similar Macro F1 score of about 0.55 for a similar problem in the scope of the NPS classification task \cite{Rallis2020, Markoulidakis2020}. The observed low statistical accuracy of our baseline and previous approaches suggests that there should be additional attributes not captured in the sentiment of the whole interaction or surveys that explain the recommendation behavior of the users, which is the problem we are trying to tackle here.


\begin{table}[]
\centering
\caption{XGBoost Classification results in terms of Area Under the Curve (AUC), Kolmogorov-Smirnov (KS) and Macro F1 score for our three experiments on the NPS binary classification task. As such, we have depicted the abbreviations for each experiment as: baseline \textbf{(B)}, our message-wise sentiment evolution analysis including \textit{passive} users \textbf{(B + LW)}, and line-wise sentiment evolution ignoring \textit{passive} users \textbf{(B+LW\{NP\})}.}
\begin{tabular}{c|ccc|}
\cline{2-4}
\textbf{}                                   & \multicolumn{3}{c|}{\textbf{XGBoost}}                            \\ \hline
\multicolumn{1}{|c|}{\textbf{Experiment}} & \multicolumn{1}{c|}{\textbf{AUC}} & \multicolumn{1}{c|}{\textbf{KS}} & \textbf{Macro F1} \\ \hline
\multicolumn{1}{|c|}{\textbf{B}}            & \multicolumn{1}{c|}{0.5513} & \multicolumn{1}{c|}{0.0801} & 0.54 \\ \hline
\multicolumn{1}{|c|}{\textbf{B + LW}}       & \multicolumn{1}{c|}{0.6199} & \multicolumn{1}{c|}{0.1843} & \textbf{0.58} \\ \hline
\multicolumn{1}{|c|}{\textbf{B + LW\{NP\}}} & \multicolumn{1}{c|}{\textbf{0.6455}} & \multicolumn{1}{c|}{\textbf{0.2389}} & \textbf{0.58} \\ \hline
\end{tabular}%
\label{tab:table1}
\end{table}

The results for our \textbf{(B + LW)} experiment are significantly better than those exhibited by our baseline \textbf{(B)} and also than those reported in previous works, in terms of the AUC, KS, and F1 scores. This suggests that the message-wise sentiment evolution analysis and derived features improve the predictive power of the classification algorithm at hand, having performance gains in terms of the AUC of about 10-14\% in any case. Moreover, when observing the validation metrics for the \textbf{(B+LW\{NP\})} experiment, it is evident that the AUC and the KS metrics further increase, so that the model is separating better the \textit{detractors} from \textit{promoters}, even when the problem had a higher class imbalance. This might be because the input space for \textit{passive} users is more diverse and sparse than for both detractors or promoters. Thus, suggesting that there might not be clear patterns to learn from the \textit{passives} population so that extensive inspection in that regard is encouraged in future works.


\begin{figure}
    \centering
    \includegraphics[width=0.8\textwidth]{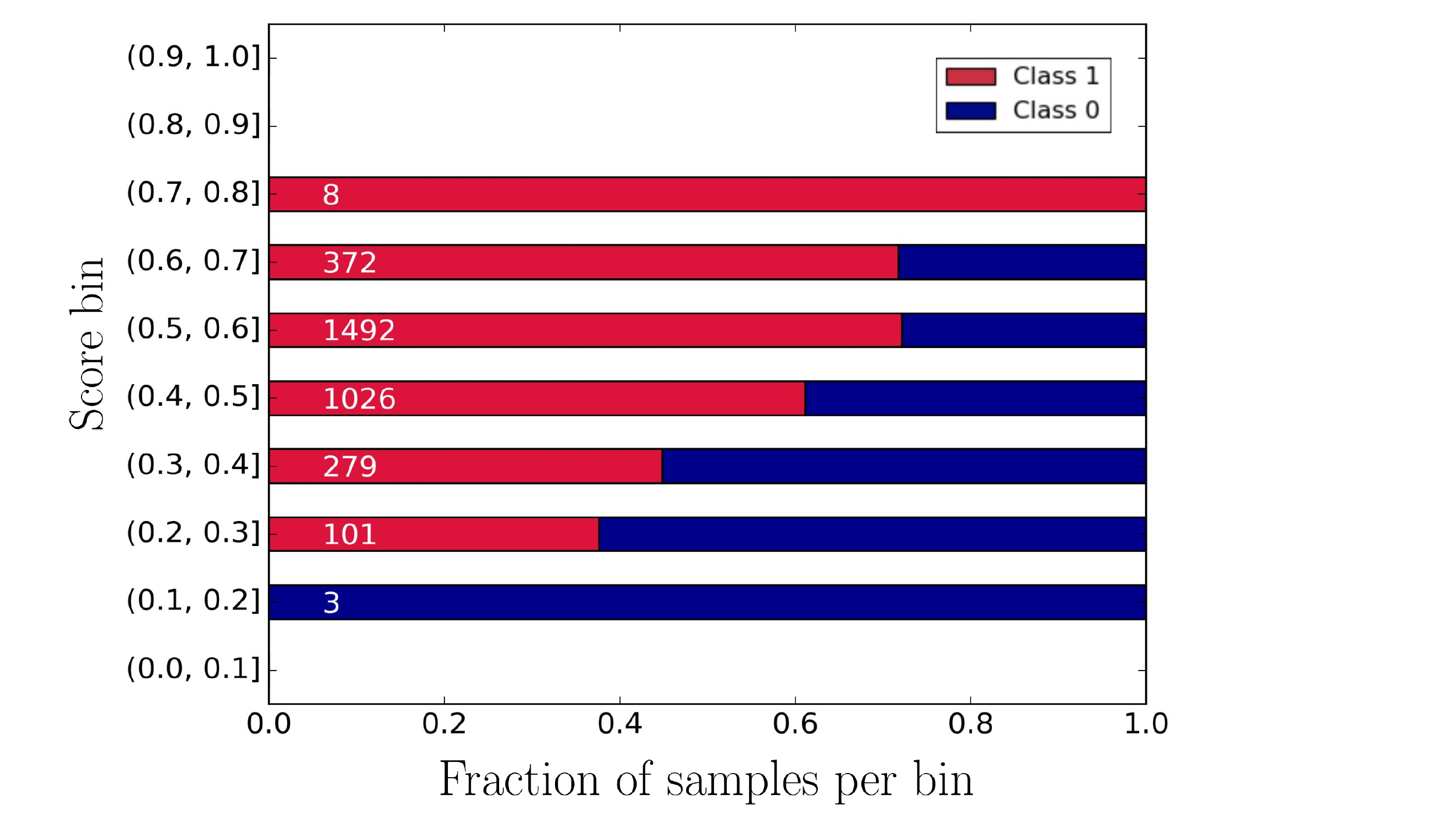}
    \caption{Relative distribution of number of samples belonging to each respective \textit{promoters}/\textit{non-promoters} class per score bin. The score represents the probability of each sample belonging to the \textit{promoters class} (Class 1) according to the trained XGBoost model for our \textbf{(B+LW)} experiment. Numbers inside each bin represent the number of samples classified per each score interval.}
    \label{fig:Scorecard}
\end{figure}


Figure \ref{fig:Scorecard} show a plot of the scorecard for our XGBoost model for the \textbf{(B+LW)} experiment. It can be seen that overall, the relative number of users classified as \textit{non-promoters}  does increase monotonically as the score decreases, which is a good indicator in this case. The score represents the probability of a sample belonging to the \textit{promoters} class. In that regard, the model is capable of separate between each class, though it is not entirely sure about none, given that the bins with scores higher (lower) than 0.8 (0.1) are empty. This can be due to a fundamental selection bias, as users contact CS only when they have issues with the product or service. Correspondingly, their recommendation behavior might be biased and lowered by itself considering the inherent nature of the CS interactions. However, the here presented results, are significantly better to those obtained in previous works \cite{Markoulidakis2020} considering explicit quantitative service ratings, for instance.

\begin{figure}
    \centering
    \includegraphics[width=0.8\textwidth]{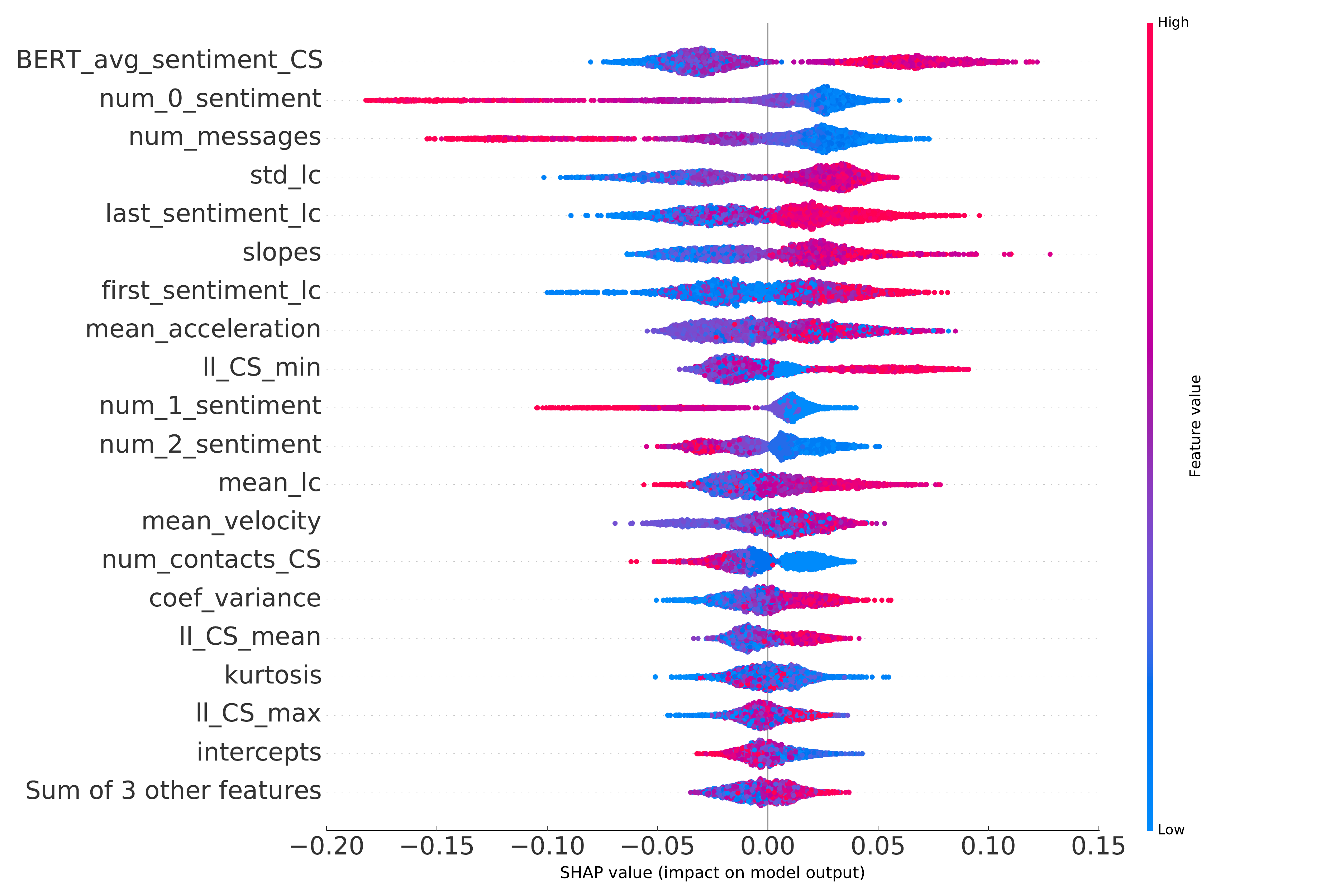}
    \caption{SHapley Additive exPlanations (SHAP) values for individual features of the line-wise sentiment analysis considering \textit{passive} users (B + LW) experiment.}
    \label{fig:SHAP}
\end{figure}

In addition, we use the SHapley Additive exPlanations (SHAP) \cite{Lundberg2017} algorithm to obtain the feature-level interpretability of our ensemble model. SHAP values can tell us the extent and relationship to which each feature in a model has contributed to the final prediction. Figure \ref{fig:SHAP} shows the SHAP values for our \textbf{(B + LW)} experiment. These values are in very good agreement with a rather intuitive reasoning, as well as previous similar studies based on text reviews \cite{Chatterjee2019}. For instance, we can see that a higher number of messages from a customer to the CS center leads to a lower value on the classification task, i.e., a lower probability of being a \textit{promoter} user. This is very sensible given that if a single client has sent a very large number of messages, it might indicate angriness or frustration, then leading to a rather poor customer experience, and thus a low recommendation decision. A similar interpretation can be extended to the number of messages having the lowest possible sentiment in our discrete sentiment scale, as well as the other corresponding features.


Furthermore, when interpreting variables such as the last sentiment of the longest conversation, the slope of the linear fitting (Fig. \ref{fig:line-wise messages}), or the average sentiment over all CS conversations, we can see that the larger these values, the more likely a person to recommend the service is. This is very relevant given that a high slope value, means that a given conversation started rather neutral/negative but ended rather positive, and this might be associated with low customer frustration and then better recommendation behavior overall, thus validating our sensible hypothesis stated in Section \ref{section:methodology}.

Very relevantly, it can be observed that the concavity analysis demonstrated that rather low acceleration values (i.e., low mean value of the numerical second derivative) leads to rather low SHAP values, whereas rather convex-like sentiment series (high acceleration values) leads to higher SHAP values. This is also in very good agreement with intuitive reasoning, as for a twice-differentiable function, if its acceleration is positive, then the line-wise series is concave upward. Likewise, if the second derivative is negative, then the graph is concave downward. Therefore, the more convex-like the line-wise sentiment series is, the more likely the user to be a \textit{promoter} is. For instance, the overall evolution of a convex-like conversation would be, in general, a conversation starting very gentle and positive, then swinging through a rather neutral-negative stage, then ending rather positively. Therefore such a convex-like pattern is also associated with better recommendation behavior, whereas concave-like shape results generally in the opposite outcome.

The more noisy and volatile a series is, the more likely the user is to be a \textit{promoter}, as confirmed by the coefficient of variation. This might be due to the fact that non-\textit{promoter} users might tend to have a more negatively biased behavior, causing also the overall standard deviation of the distribution to be small. Furthermore, for the sum of other features in the last row of Fig. \ref{fig:SHAP} we can see that their overall aggregated SHAP values do not have strong predicting power, yet can further segment our target population.


In general, the here presented results represent the first analysis and assessment of the actual predictive powers of the message-wise sentiment evolution throughout chat-based CS interactions on the recommendation behavior of individual users. As such, this work represents a highly flexible and interpretable framework for proactive intervention of \textit{promoter}/\textit{non-promoter users} which is particularly relevant, given that in our case-study the NPS survey is sent periodically once every few months. Therefore, the framework could be used in other classification tasks regarding other business-driven metrics or customer ratings based on chat-based interactions in any particular domain.

Finally, we would like to discuss that even though we do not have a clear baseline model present in the literature to which we can fairly compare against, we do observe that our baseline is comparable in performance to different available studies in the field \cite{Siering2018, Markoulidakis2020, Chatterjee2019, Auguste2018}. Generally, the lack of baselines or benchmark comparisons are grounded in the fact that there are no publicly available datasets containing chat-based conversations with NPS annotations obtained directly from the users; as these datasets typically belong to private companies, highly regulated by the authorities, as in our study case. However, there is no foreseeable fact that would prevent the here presented results and employed ML techniques to be generally extensible to any industry domain.

\section{Conclusions}
\label{section:conclusions}

In this work, we propose and evaluate our framework on the NPS classification problem in the field of CS interactions, showing its value, flexibility and interpretability in a real-world case study, where rating annotations were provided directly by the users in corresponding surveys. Our results show that it is possible to predict the recommendation decision of users based on dynamic sentiment classification of chat-based data sources employing transformer-based methods. Results show performance gains of about 10-14\% obtained when considering a sentiment evolution analysis versus a purely aggregated, review-based, sentiment classification. Moreover, our explainable features explicitly allowed us to draw important and intuitive insights regarding the complex relations that do arise between the recommendation behavior of a given user and the sentiment evolution of their CS messages. More importantly, we here present a framework that can be easily extended to other prediction tasks in any business environment, considering any customer ratings of interest. We therefore hope that this work could spark interest among data-driven companies, and will inspire other works in this research area, where inherent attributes related to the evolution of conversations could be exploited to further develop state-of-the-art techniques and applications.


\section*{Acknowledgements}
We would like to thank Luisa F. Roa, Ana M. Quintero, Jaime Acevedo, V. Maya, and R. Saavedra for their insightful ideas and valuable discussions during the development of this research project.

\end{document}